\pdfoutput=1

\documentclass[11pt]{article}

\usepackage{naacl2021}

\usepackage{times}
\usepackage{latexsym}
\usepackage{multirow}
\usepackage{booktabs}
\usepackage{graphicx}
\usepackage{float}
\usepackage{enumitem}
\usepackage{color,soul}
\usepackage{todonotes}

\setlist{nosep}

\usepackage[T1]{fontenc}

\usepackage[utf8]{inputenc}

\usepackage{microtype}

\title{Classifying Long Clinical Documents with Pre-trained Transformers}

\author{
    \begin{tabular}{c@{\hskip 1.5cm}c@{\hskip 1.5cm}c}
    Xin Su\textsuperscript{1} & Timothy Miller\textsuperscript{2} & Xiyu Ding\textsuperscript{2} \\
    \rule{0pt}{.5cm} Majid Afshar\textsuperscript{3} & Dmitriy Dligach\textsuperscript{4}\\
    \end{tabular} \vspace*{.35cm} \\
    \textsuperscript{1}University of Arizona \\
    {\footnotesize\tt xinsu@email.arizona.edu} \\
    \textsuperscript{2}Boston Children’s Hospital and Harvard Medical School \\
    {\footnotesize\tt timothy.miller@childrens.harvard.edu} \\
    {\footnotesize\tt xiyu\_ding@hsph.harvard.edu} \\
    \textsuperscript{3}University of Wisconsin–Madison \\
    {\footnotesize\tt mafshar@medicine.wisc.edu} \\
    \textsuperscript{4}Loyola University Chicago \\
    {\footnotesize\tt dd@cs.luc.edu}
}

\begin{document}
\maketitle
\begin{abstract}
Automatic phenotyping is a task of identifying cohorts of patients that match a predefined set of criteria. Phenotyping typically involves classifying long clinical documents that contain thousands of tokens. At the same time, recent state-of-art transformer-based pre-trained language models limit the input to a few hundred tokens (e.g. 512 tokens for BERT). We evaluate several strategies for incorporating pre-trained sentence encoders into document-level representations of clinical text, and find that hierarchical transformers without pre-training are competitive with task pre-trained models.
\end{abstract}

\section{Introduction}

Text encoding is a key element of modern natural language processing (NLP) and a prerequisite for applying machine learning algorithms to text. Numerous text encoding algorithms have been proposed with transformer-based~\citep{NIPS2017_7181} pre-trained methods advancing the state of the art in the field to unprecedented levels. However, due to hardware limitations and computational complexity (quadratic in the length of the input for transformers), most models are limited to a maximum length of the input text, usually to several hundred tokens (e.g. 512 tokens for BERT and RoBERTa~\citep{devlin-etal-2019-bert, liu2019roberta}).

These length limitations restrict the usefulness of pre-trained models for many practical clinical tasks. Computable phenotyping, the task of identifying cohorts of patients that match a predefined set of criteria from the electronic health record (EHR), is one such important task in medical informatics. Phenotyping is typically performed at the encounter- or patient-level, and with the large amount of text generated in a clinical encounter, the average length of a document (a unit of classification in phenotyping) is often in the thousands if not tens of thousands of tokens. For example, the average number of words in a MIMIC III~\citep{johnson2016mimic} encounter is 8,131.

The large size of clinical documents makes it difficult to directly use current pretrained text encoders for computable phenotyping. The truncation approach~\citep{sun2019fine}, where only the first (or last) 512 tokens are used is obviously unsatisfying due to significant information loss in the input representations. \citet{adhikari2019docbert} distills the BERT model to a much simpler neural model to reduce the number of parameters. There have been several recent attempts \citep{Kitaev2020Reformer:, Beltagy2020Longformer} to reduce the time and space complexity of the self-attention mechanism in the transformer; however, these methods still cannot represent long enough texts, and for best performance require pretraining from scratch.

This work explores approaches for encoding long documents that leverage powerful pre-trained encoders to generate representations of short sequences, aggregating these chunk representations into document-level representations. We chose DistilBERT~\citep{sanh2019DistilBERT} as our chunk encoder for its computational efficiency. Our contributions are:
\begin{enumerate}
    \item We propose a simple framework that can integrate and apply any existing transformer encoder-based pre-trained model to long clinical text classification. The framework does not require any architectural changes to the pre-trained encoder and does not require any complex data pre-processing.
    \item We show that our framework can improve the performance of the DistilBERT model across 16 long clinical text classification tasks. Without any task specific pre-training, the model achieves similar performance to the task specific pre-trained model.
    \item We systematically investigate different strategies for using the representations from the DistilBERT model. We find the best performance from the strategy of using transformer-based chunk pooling along with end-to-end training.
\end{enumerate}

Our primary finding is that a hierarchical transformer approach that leverages even smaller pretrained sentence encoders is competitive with a more complex model pretrained on task-related data. This suggests that this is a promising direction for future work in representing long clinical texts. Our code will be made publicly available upon publication.

\begin{figure}[H]
\centering
\includegraphics[scale=0.30]{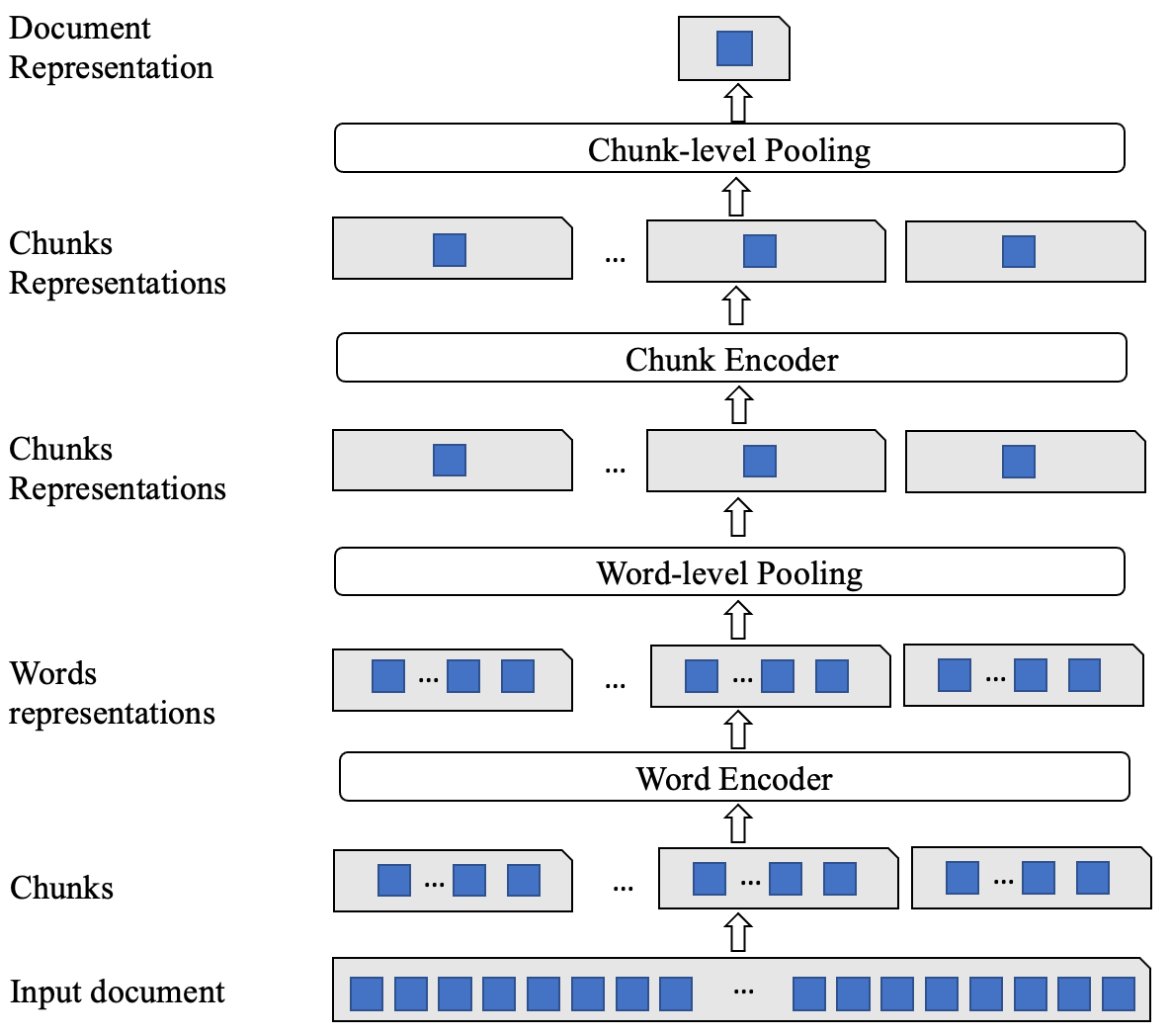}
\caption{The architecture of our hierarchical framework.}
\label{fig:Model-Struct}
\end{figure}

\section{Methods}

We develop a hierarchical framework for encoding long text documents. To model a long document, we begin by dividing the document into chunks of equal length and insert the \verb|[CLS]| token at the beginning of each chunk.\footnote{We have explored using sentence segmentation to ensure chunks do not break up sentences, but saw little difference in performance.} We deliberately divide the document into equal-length chunks to avoid the use of complex sentence segmentation algorithms (often tied to idiosyncrasies of a particular institution) and the extra padding work it would bring.  We choose the length of each chunk to be short enough to be handled by pre-trained DistilBERT (typically several hundred tokens). We utilize a word-level encoder and pooling to form chunk representations, and chunk-level encoder and pooling to form document representations. The architecture of our framework is shown in Figure \ref{fig:Model-Struct}. 

\subsection{Word Encoder and Word-level Pooling}

We used DistilBERT as our word-level encoder and iteratively run it on each chunk to obtain the contextual representation of each token in the chunk. We explore two different word-level pooling methods to obtain the chunk representations.

The first method, \textbf{[CLS] Pooling}, simply uses the contextual representation of the \verb|[CLS]| token as each chunk's representation. We also explore a \textbf{weighted sum} approach, in which we first extract a layer representation by averaging the hidden state of all the tokens from each of 6 layers of the pretrained DistilBERT encoder. We calculate the weighted sum of these 6 different representations to obtain the final chunk representation: $R_{final} = \sum_{i=1}^{6} w_ih_i$, where $R_{final}$ is the final chunk representation, $h_i$ is the hidden state from the $i_{th}$ layer of the DistilBERT, and $w_i \in  R$ is the weight associated with each hidden state (a trainable parameter).

\subsection{Chunk Encoder and Chunk-level Pooling}

To aggregate chunks into documents, we explore several forms of pooling: (1) a transformer encoder, (2) an LSTM, (3) a CNN, and (4) average pooling. The representation of a document is tied to the type of encoder and pooling mechanism.

For the transformer encoder, we add an empty chunk at the beginning of the chunk sequence that is used as the document representation to the classifier, analogous to the \verb|[CLS]| token commonly used in that way for sentence representations (and used here as the chunk representation). The transformer layer we use is similar to the transformer layer in the DistilBERT model. Each of its layers is composed of two sub-layers, namely the multi-head self-attention layer and the position-wise feed-forward neural network. However, we use a smaller transformer for the top of the hierarchy, to reduce the number of parameters that we need to learn, since we are training on a supervised task without pre-training. Positional information can be injected into the model by adding position embeddings~\citep{gehring2017convolutional} to chunk representations. To do this, we re-use the pretrained position embedding layer from DistilBERT. 

The LSTM encoder uses the last hidden state of an LSTM learned over the sequence of chunk representations as a document representation. In the CNN encoder, We use the fixed-length representation obtained by applying a convolution and max-pooling operations on the sequences of chunk representations. Finally, we experiment with simply averaging the chunk representations to obtain a document representation. 

In all variants of the model, the resulting hierarchical model can be fine-tuned end-to-end. Alternatively, the word-level encoder can be `frozen' and used as a feature extractor. We evaluate both options.

\begin{table*}
\small
\centering
\begin{tabular}{llllll}
\toprule
\multicolumn{3}{l}{\textbf{Fine-tuning DistilBERT}} \\
\midrule
\textbf{\#} & \textbf{WP} & \textbf{CE} & \textbf{CP} & \textbf{F (i2b2)} & \textbf{ROC (Injury)} \\
\midrule
1 & [CLS] & Transformer & First Chunk & \textbf{0.7434} & \textbf{0.8860} \\
2 & [CLS] & Transformer w/ Position & First Chunk  &0.7154 & 0.8652 \\
3 & [CLS] & LSTM & Last Hidden & 0.3747 & 0.5700 \\
4 & [CLS] & CNN & Max pooling & 0.6822 & 0.8700 \\
5 & [CLS] & Average Pooling & \ & 0.5615 & 0.8129 \\
\midrule
\multicolumn{3}{l}{\textbf{DistilBERT as Feature Extractor}}\\
\midrule
6 & Weighted Sum & Transformer & First Chunk  &0.4483 & 0.6237\\
7 & Weighted Sum & Transformer w/ Position & First Chunk  &0.4484 & 0.6238\\
8 & Weighted Sum & LSTM & Last Hidden & 0.3701 & 0.3930 \\
9 & Weighted Sum & CNN & Max pooling & 0.4678 & 0.6935 \\
10 & Weighted Sum & Average Pooling & \ & 0.4207 & 0.3806 \\
\midrule
\multicolumn{3}{l}{\textbf{Models for comparison}} \\
\midrule
11 & \multicolumn{3}{l}{SVM, no pretraining~\cite{si2020patient}} &  0.6763 &\\
12 & \multicolumn{3}{l}{SVM, with two-level HAN features~\cite{si2020patient}} &  \textbf{0.7525} & \\
13 & \multicolumn{4}{l}{SVM, no pretraining} & \textbf{0.9101} \\
14 & \multicolumn{3}{l}{DistilBERT} &  0.6000 & \\
\bottomrule
\end{tabular}
\caption{\label{models-performance}
Summary of the performance of the models on i2b2 Obesity and Injury Severity datasets. WP is world-level pooling. CE is chunk-level encoder. CP is chunk-level pooling. F is average macro F1 scores across the 16 diseases in i2b2 Obesity dataset.
}
\end{table*}

\section{Data}

We evaluate our framework on the publicly available i2b2 (Informatics for Integrating Biology to the Bedside) 2008 Obesity Challenge~\citep{uzuner2009recognizing} dataset. The i2b2 dataset is available at DBMI Data Portal\footnote{\url{https://portal.dbmi.hms.harvard.edu/}} for general research purposes. To validate our framework in a real-world clinical environment, we also test our model on the Injury Severity dataset from our medical center. 

\subsection{i2b2 2008 Obesity Challenge dataset}

The i2b2 2008 Obesity Challenge dataset is composed of discharge summaries of 1,237 patients from Partners HealthCare Research Patient Repository. The average length of a document is 1836 word pieces. There are 1230 documents with more than 512 word pieces. We test our model on the more challenging intuitive task from the challenge. The task is to identify whether the patient has obesity and 15 other concurrent diseases based on the provided discharge summary (a total of 16 classification tasks). For each disease, the discharge summary is labeled as present, absent or questionable. Following the primary metric in the challenge, we use macro-averaged F-score as our evaluation metric. 

\subsection{Injury Severity dataset}

Our in-house Injury Severity dataset (IRB approved) consists of clinical notes from the EHR captured during the first hour of arrival to the trauma center at a large academic medical center. Severe chest injury (positive sample) was labeled using a thorax abbreviated injury score (AIS) cutoff for serious injury (AIS$>$2). The AIS scores were labeled by trauma registry specialists who have been credentialed and certified through the Registrar Certifying Board of the American Trauma Society, which is the gold standard for quality reporting. We split the training set and test set according to the ratio of $80\%:20\%$. There are 361 positive samples and 5152 negative samples in the training set, and 107 positive samples and 1271 negative samples in the test set. The average length of a document is 1819 word pieces. There are 5210 documents with more than 512 word pieces. In order to speed up training and deal with the problem of unbalanced dataset, we down-sample the training set so that the positive and negative samples in the training set have the same number (a total of 722 samples). Area under the ROC curve is used as the primary evaluation metric.

\section{Experiments}

\subsection{Comparisons}

For the i2b2 Obesity dataset, we evaluate several comparators. Plain DistilBERT uses the first 512 tokens as input. The two SVM models from \citet{si2020patient} use bag-of-words features and use domain specific pre-trained two-level Hierarchical Attention Network (HAN)~\citep{yang-etal-2016-hierarchical} features respectively. The two-level RNN based HAN model from \citet{si2020patient} was pretrained on the MIMIC III dataset using a task-specific source of supervision (billing codes related to i2b2 task labels) and used a segmentation algorithm to preprocess the input. The patient's discharge summary is passed to the pretrained RNN-based HAN model with frozen weights to obtain a dense patient representation of 100 dimensions. An SVM model is then trained using these patient representations as features. For the Injury Severity dataset, we use a linear SVM model with bag-of-words features as the baseline.

\subsection{Training}

We remove all punctuation from the input text and use WordPiece tokenizer~\citep{wu2016google} to segment the input text into tokens. We split each input document into chunks containing 202 tokens. All models are trained by Adam optimization algorithm with a linear learning rate warm-up. We report the settings of the encoders in table \ref{transformer-setting}, \ref{CNN-setting} and \ref{LSTM-setting} in the appendix. We also report the hyperparameters used in training in table \ref{models-hyperpatameters-1} and \ref{models-hyperpatameters-2} in appendix.

\subsection{Results and Discussion}

We report all models' performance in Table~\ref{models-performance}. For i2b2 Obesity challenge dataset, we report the average Macro F1 scores across the 16 classification tasks. For injury severity dataset, we report the AUC ROC scores.

On the i2b2 Obesity challenge dataset, our best performing model is fine-tuned DistilBERT model with transformer-based chunk aggregation (row 1 in table \ref{models-performance}). It significantly outperforms the DistilBERT model by +.1434 F1 scores (row 1 vs row 14 in table \ref{models-performance}) and outperforms the SVM model using bag-of-word features by +.0671 F1 scores (row 1 vs row 11 in table \ref{models-performance}). This means our framework 
can effectively apply the DistilBERT model to long clinical document classification. Our best model achieves similar performance to the pretrained HAN model from \citet{si2020patient} (row 1 vs row 12 in table \ref{models-performance}). Importantly, HAN relied on domain-specific pretraining that involved hand-selecting pretraining targets (ICD codes) relevant for the obesity classification task. In contrast, our model used an off-the-shelf model pre-trained on general domain data without any additional pre-training and, unlike HAN, did not require complex data preprocessing.

Our best performing model achieves reasonable performance on our in-house Injury Severity dataset, but does not outperform the baseline model (row 1 vs row 13 in table \ref{models-performance}). This may be an indication that this task relies on a relatively small set of key words and does not require complex understanding of clinical language.

If we add position information to our model (row 2 in table \ref{models-performance}), the performance on the i2b2 Obesity dataset drops (row 1 vs row 2 in table \ref{models-performance}), indicating that chunk position information was not important for this task.

To investigate the effect of chunk-level encoders, we change the chunk-level encoders to LSTM, CNN and average pooling resulting in row 3-5 and 8-10 in table \ref{models-performance}. The worst-performing chunk-level encoder is LSTM (rows 3 and 8 in table \ref{models-performance}). It indicates that LSTM may not be suitable for combining chunk representations from transformers. We leave an investigation into the exact reasons for such low performance for future work. The performance of the CNN encoder is better than the baseline (row 4 vs row 11 in table \ref{models-performance}) and significantly better than average pooling (row 4 vs row 5 in table \ref{models-performance}).

Finally, the performance of all feature extraction based models (rows 6-10 in table \ref{models-performance}) on both i2b2 Obesity challenge and Injury Severity datasets are significantly lower than that of the fine-tuned and baseline models. This suggests that fine-tuning DistilBERT is critical for aggregating chunk representations effectively.

\section{Conclusion}

We investigated several approaches for aggregating representations of short text fragments into document-level representations. Our main finding is that it is possible to effectively combine the representations obtained from an off-the-shelf pretrained text encoder into a representation of a long document. We investigated a number of pooling mechanisms including Transformer encoder, LSTM, CNN, and averaging. It appears that Transformer encoder-based pooling is the best approach for aggregating the representations of short fragments based on our experiments with two datasets. Finally, end-to-end training is significantly more effective than using pretrained encoders as feature extractors.


\bibliography{anthology,custom}
\bibliographystyle{acl_natbib}

\appendix

\section{Appendix}
\label{sec:appendix}

\begin{table}[H]
\centering
\begin{tabular}{ll}
\toprule
\textbf{Settings} & \textbf{Value} \\
\midrule
\# Layers &  2 \\
Hidden Size & 768 \\
Inner FF Size & 2048 \\
\# Attention Heads & 8 \\
Attention Key Size & 8 \\
Attention Value Size & 96 \\
\bottomrule
\end{tabular}
\caption{The settings of transformer encoder. The FF is feed forward layer.}
\label{transformer-setting}
\end{table}

\begin{table}[H]
\centering
\begin{tabular}{ll}
\toprule
\textbf{Settings} & \textbf{Value} \\
\midrule
\# Layers &  1 \\
Hidden Size & 768 \\
Kernel Size & 3 \\
\bottomrule
\end{tabular}
\caption{The settings of CNN encoder.}
\label{CNN-setting}
\end{table}

\begin{table}[H]
\centering
\begin{tabular}{ll}
\toprule
\textbf{Settings} & \textbf{Value} \\
\midrule
\# Layers &  1 \\
Hidden Size & 768\\
Direction & Unidirectional \\
\bottomrule
\end{tabular}
\caption{The settings of LSTM encoder.}
\label{LSTM-setting}
\end{table}

\begin{table}[H]
\centering
\begin{tabular}{ll}
\toprule
\textbf{Hyperparameter} & \textbf{Value} \\
\midrule
Learning Rate &  3e-5 \\
Batch size & 16 \\
Gradient Accumulation Steps & 2 \\
Epochs & 40 \\
Warm-up Steps & 150 \\
\bottomrule
\end{tabular}
\caption{The hyperparameters for training fine-tuned models.}
\label{models-hyperpatameters-1}
\end{table}

\begin{table}[H]
\centering
\begin{tabular}{ll}
\toprule
\textbf{Hyperparameter} & \textbf{Value} \\
\midrule
Learning Rate &  3e-5 \\
Batch size & 32 \\
Gradient Accumulation Steps & 1 \\
Epochs & 20 \\
Warm-up Steps & 40 \\
Warm-up Steps (injury) & 20 \\
\bottomrule
\end{tabular}
\caption{The hyperparameters for training feature extraction models.}
\label{models-hyperpatameters-2}
\end{table}

\end{document}